\documentclass[10pt,twocolumn,letterpaper]{article}

\usepackage{subfigure}
\usepackage{iccv}
\usepackage{times}
\usepackage{epsfig}
\usepackage{graphicx}
\usepackage{amsmath}
\usepackage{amssymb}
\usepackage{multirow}
\usepackage{array}
\usepackage{epstopdf}
\usepackage{fancyhdr}
\newcommand\blfootnote[1]{%
  \begingroup
  \renewcommand\thefootnote{}\footnote{#1}%
  \addtocounter{footnote}{-1}%
  \endgroup
}

\usepackage[breaklinks=true,bookmarks=false,hyperfootnotes=false]{hyperref}

\iccvfinalcopy % *** Uncomment this line for the final submission

\def\iccvPaperID{****} % *** Enter the ICCV Paper ID here

\ificcvfinal\fi
\begin{document}
%%%%%%%%% TITLE
\title{Deep Generative Filter for Motion Deblurring}

\author{Sainandan Ramakrishnan$^{1}$, Shubham Pachori*$^{2}$, Aalok Gangopadhyay*$^{2}$, Shanmuganathan Raman$^{2}$ \\
Veermata Jijabai Technological Institute, Mumbai - 400031$^{1}$\\
Indian Institute of Technology, Gandhinagar - 382355$^{2}$\\
{\tt\small saiyanlife415@gmail.com$^{1}$, $\lbrace$shubham\_pachori, aalok, shanmuga$\rbrace^{2}$ @iitgn.ac.in}
}

\maketitle
\thispagestyle{empty}

%%%%%%%%% ABSTRACT
\begin{abstract}
Removing blur caused by camera shake in images has always been a challenging problem in computer vision literature due to its ill-posed nature. Motion blur caused due to the relative motion between the camera and the object in 3D space induces a spatially varying blurring effect over the entire image. In this paper, we propose a novel deep filter based on Generative Adversarial Network (GAN) architecture integrated with global skip connection and dense architecture in order to tackle this problem. Our model, while bypassing the process of blur kernel estimation, significantly reduces the test time which is necessary for practical applications. The experiments on the benchmark datasets prove the effectiveness of the proposed method which outperforms the state-of-the-art blind deblurring algorithms both quantitatively and qualitatively. \blfootnote{* denotes the equal contribution.}
\end{abstract}
%%%%%%%%% BODY TEXT
\section{Introduction}
Motion blur is a common problem which occurs predominantly when capturing an image using light weight devices like mobile phones. Due to the finite exposure interval and the relative motion between the capturing device and the captured object, the image obtained is often blurred. In \cite{vasiljevic2016examining}, it was shown that standard network models, trained only on high-quality images, suffer a significant degradation in performance when applied to those degraded by blur due to defocus or subject/camera motion. Thus, there is a serious need to tackle the issue of blurring in images. Blur induced due to motion in images is spatially non-uniform and the blur kernel is unknown. Due to depth variation, the segmentation boundaries of the objects and the relative motion between the camera and scene objects, estimating spatially variant non-uniform kernel is quite difficult. In this paper, we introduce a generative adversarial network (GAN) based deep learning architecture to address this challenging problem. We obtain significantly better results than the state-of-the-art algorithms proposed to solve the problem of image deblurring. 
\section{Related Work}
Most of the previous works in the literature tackle the problem of camera deshaking by modelling it as a blind deconvolution problem and using image statistics as priors or regularizers to obtain the blur kernels. While these methods have achieved great success in benchmark datasets, restrictive assumptions in their methods and algorithms limit their practical applicability. Also, most of these works in the literature have been dedicated to solve the problem of blind deconvolution assuming the blur kernel to be spatially uniform. Very few works have been proposed to solve this challenge by taking spatially varying blur kernel. To tackle the problem of non-uniform blind deblurring, previous works divide the image into smaller regions and estimate the blur kernels for each region separately \cite{harmeling2010space}. Once the kernels are obtained for each of the local regions in the image, they are then deblurred and combined using OLA (Overlap Add) method to generate the final deconvolved image. Proposed works which exploit deep learning methods first try to predict the probabilistic distribution of motion blur information in a small region of the given image and then try to utilize this blurring observation to recover the sharp image \cite{sun2015learning}. Only one work to the very best of our knowledge has attempted to directly recover the sharp image from the given blurred image \cite{nah2016deep}. However, it is computationally expensive as authors exploit multi-scale framework to obtain the deblurred image. Therefore, we aim to recover the artifact-free image directly without using the multi-scale framework. An exhaustive survey of blind deblurring algorithms can be found in \cite{lai2016comparative}.  
\begin{figure*}
\centering 
\includegraphics[width= 1\textwidth]{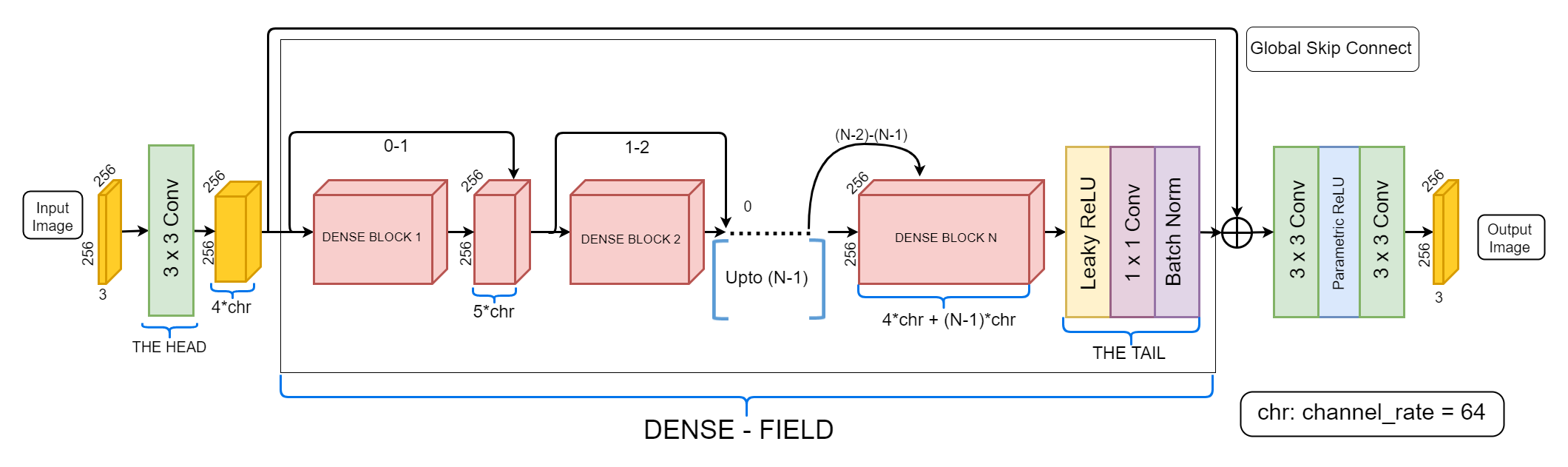}
\caption{Our Convolutional Neural Network Architecture.  \label{CNN}}
\end{figure*}
\begin{figure}
\centering 
\includegraphics[width= 0.5\textwidth]{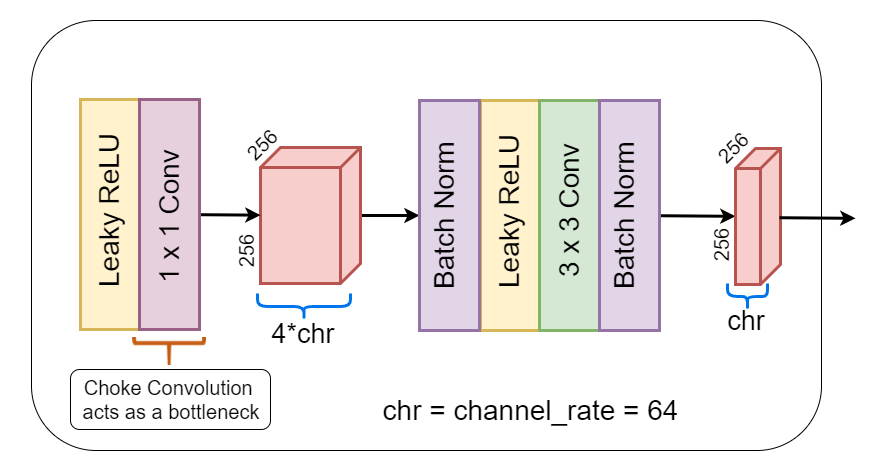}
\caption{Structure inside our Dense Block.  \label{denseblock}}
\end{figure}
\begin{figure*}[htbp]
\centering 
\subfigure[Blurry Input Image]{\includegraphics[width= 0.25\textwidth]{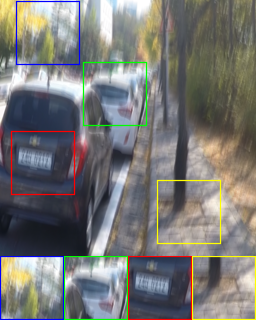}}%
\subfigure[Ground Truth]{\includegraphics[width= 0.25\textwidth]{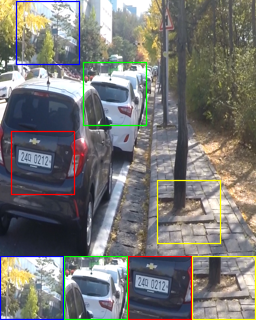}}%
\subfigure[Xu et al.\cite{xu2013unnatural}]{\includegraphics[width= 0.25\textwidth]{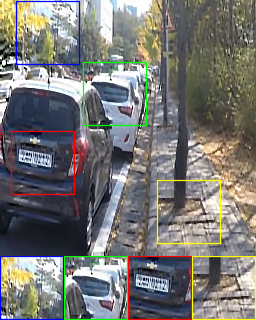}}%
\subfigure[Whyte \cite{whyte2014deblurring}]{\includegraphics[width= 0.25\textwidth]{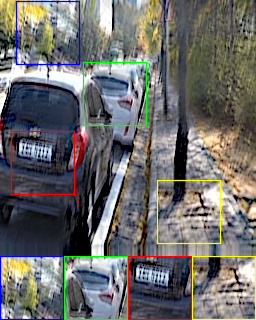}}
\subfigure[Sun et al. \cite{sun2015learning}]{\includegraphics[width= 0.25\textwidth]{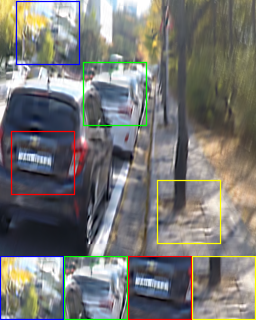}}%
\subfigure[MBMF \cite{gong2016motion}]{\includegraphics[width= 0.25\textwidth]{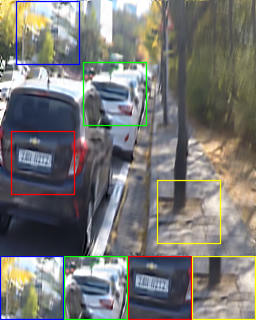}}%
\subfigure[MS-CNN \cite{nah2016deep}]{\includegraphics[width= 0.25\textwidth]{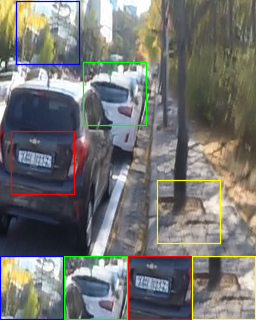}}%
\subfigure[OURS]{\includegraphics[width= 0.25\textwidth]{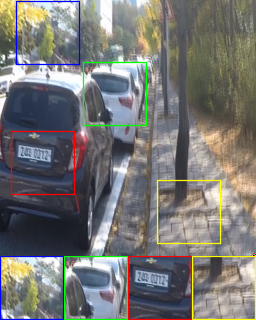}}
\caption{Comparison of deblurred images by our model and other algorihtms on one of the images taken from GoPro dataset \cite{nah2016deep}.  \label{figure1}}
\end{figure*}
\section{Proposed Method}
In our model, we enable every convolutional unit in the deep network to make independent decisions based on the entire array of lower level activations. Unlike \cite{ledig2016photo} and \cite{nah2016deep} which use residual blocks as primary workhorses through element-wise summation of lower level activations with higher level outputs, we want information from different semantic levels to flow unaltered throughout the network. To achieve this, we propose a densely connected `generative network'.
\begin{table*}[htbp]
\centering
\begin{tabular}{|c|c|c|c|c|c|c|c|c|}
\hline
Method & PSNR (dB) & SSIM & MS-SSIM & F-SIM & UIQI & IFC & VIF  \\
\hline
Ours(A) & 28.0345 & 0.8895 & 0.9678 & 0.8943 & 0.9612 & 4.0904 & 0.8691 \\
Ours(B)  & 28.5798 & 0.9090 & 0.9701 & 0.9132 & 0.9683 & 4.2458 & 0.8749 \\
\hline
Ours(final) & \textbf{28.9423} & \textbf{0.9220} & \textbf{0.9720} & \textbf{0.9248} & \textbf{0.9741} & \textbf{4.9455} & \textbf{0.8853}\\
\hline
\end{tabular}
\caption{GoPRO Test Dataset (Ablation study on generative dense-net architecture), Ours(A): Residuals at extremes, dense in the middle,
Ours(B): Dense across extremes, successive residuals in the middle. \label{Table1}}
\end{table*}

\begin{table}[htbp]
\centering
\begin{tabular}{|c|c|c|c|c|c|c|c|c|}
\hline
Method & MBMF \cite{gong2016motion} & MS-CNN \cite{nah2016deep} & OURS  \\
\hline
Time  & 0.72 sec & 2.2 sec & \textbf{0.3} sec \\
\hline
\end{tabular}
\caption{Average time to deblur the input image of size $256\times256\times3$. \label{Table6}}
\end{table}

\begin{table*}[htbp]
\centering
\begin{tabular}{|c|c|c|c|c|c|c|c|c|}
\hline
Method & PSNR (dB) & SSIM & MS-SSIM & F-SIM & UIQI & IFC & VIF  \\
\hline
ResGAN \cite{ledig2016photo} & 24.3460 & 0.7678 & 0.8697 & 0.8352 & 0.9715 & 2.1568 & 0.7043 \\
Pix2Pix \cite{isola2016image} & 24.5987 & 0.7692 & 0.8680 & 0.8379 & 0.9675 & 2.0354 & 0.6992 \\
Ours(1) & 24.5281 & 0.7625 & 0.8551 & 0.8113 & 0.9421 & 1.9805 & 0.6835 \\
Ours(2)  & 24.5412 & 0.7656 & 0.8602 & 0.8310 & 0.9455 & 2.0051 & 0.6981 \\
Ours(3) & 24.6991 & 0.7677 & 0.8681 & 0.8354 & 0.9532 & 2.1143 & 0.7038 \\
Ours(4)  & 25.4897 & 0.7718 & 0.8694 & 0.8417 & 0.9679 & 2.3875 & 0.7315 \\
Ours(5) & 26.8134 & 0.8081 & 0.8840 & 0.8733 & 0.9758 & 2.5892 & 0.7581 \\
\hline
Ours(final) & \textbf{27.0812} & \textbf{0.8362} & \textbf{0.9112} & \textbf{0.8936} & \textbf{0.9778} & \textbf{2.9348} & \textbf{0.7740}\\
\hline
\end{tabular}
\caption{Quantitative Comparison of Progressive Model with Benchmarks on Synthetically blurred Places Dataset. Ours(1): Without Perceptual Loss, Ours(2): Without GAN (with (1)), Ours(3): Without conditional GAN (with(1,2)), Ours(4): Without global skip connection (with(1,3)), Ours(5): Without dilated convolution (with(1,3,4)) and Ours(final): with(1,3,4,5). \label{Table2}}
\end{table*}
\begin{table*}[htbp]
\centering
\begin{tabular}{|c|c|c|c|c|c|c|c|c|}
\hline
Method & PSNR (dB) & SSIM & MS-SSIM & F-SIM & UIQI & IFC & VIF & Norm-NR  \\
\hline
Xu et al.\cite{xu2013unnatural}  & 25.1858 & 0.8960 & 0.9614 & 0.9081 & 0.9527 & 4.1811 & 0.8644 & 0.9570  \\
Sun et al. \cite{sun2015learning} & 24.6890 & 0.8561 & 0.9308 & 0.8691 & 0.9427 & 4.1132 & 0.8430 & 0.9532 \\
MBMF \cite{gong2016motion} & 27.1989 & 0.9082 & 0.9617 & 0.9138 & 0.9450 & 4.2032 & 0.8699 & 0.9581 \\
MS-CNN \cite{nah2016deep} & 28.4496 & 0.9165 & \textbf{0.9729} & 0.9073 & 0.9693 & 4.1969 & 0.8752 & \textbf{0.9657} \\
\hline
Ours (final) & \textbf{28.9423} & \textbf{0.9220} & 0.9720 & \textbf{0.9248} & \textbf{0.9741} & \textbf{4.9455} & \textbf{0.8853} & 0.9642\\
\hline
\end{tabular}
\caption{Quantitative Comparison of our method with other state-of-the-art blind deblurring algorithms on GoPro Dataset. \label{Table3}}
\end{table*}
\begin{table*}[htbp]
\centering
\begin{tabular}{|c|c|c|c|c|c|c|c|c|}
\hline
Method & PSNR (dB) & SSIM & MS-SSIM & F-SIM & UIQI & IFC & VIF & Norm-NR  \\
\hline
Xu et al.\cite{xu2013unnatural} & 25.95 & 0.7474 & \textbf{0.8358} & 0.8309 & \textbf{0.9563} & 2.4140 & 0.7478 & \textbf{0.9271}  \\
Whyte \cite{whyte2014deblurring} & 24.41 & 0.7312 & 0.8033 & 0.8293 & 0.9524 & 2.3910 & 0.7298 & 0.9103 \\
Sun et al.\cite{sun2015learning} & 24.58 & 0.7379 & 0.8059 & 0.8255 & 0.9393 & 2.3897 & 0.7303 & 0.9087 \\
MBMF \cite{gong2016motion} & 25.87 & 0.7420 & 0.8157 & 0.8136 & 0.9418 & 2.4385 & 0.7398 & 0.9201 \\
MS-CNN \cite{nah2016deep} & 26.79 & 0.7572 & 0.8168 & 0.8311 & 0.9535 & 2.4211 & 0.7317 & 0.9128 \\
\hline
Ours (final) & \textbf{27.23} & \textbf{0.7651} & 0.8217 & \textbf{0.8712} & 0.9538 & \textbf{2.6158} & \textbf{0.7597} & 0.9214\\
\hline
\end{tabular}
\caption{Quantitative Comparison of our method with other state-of-the-art blind deblurring algorithms on Lai Dataset.\label{Table4}}
\end{table*}
\begin{table*}[htbp]
\centering
\begin{tabular}{|c|c|c|c|c|c|c|c|c|}
\hline
Method & PSNR (dB) & SSIM & MS-SSIM & F-SIM & UIQI & IFC & VIF & Norm-NR  \\
\hline
Xu et al.\cite{xu2013unnatural}  & \textbf{27.47} & 0.7506 & 0.8115 & \textbf{0.8810} & 0.9642 & 2.5025 & 0.7698 & 0.9309  \\
Whyte \cite{whyte2014deblurring} & 27.03 & 0.7467 & 0.8091 & 0.8802 & 0.9589 & 2.4556 & 0.7632 & 0.9287 \\
Sun et al. \cite{sun2015learning} & 25.12 & 0.7281 & 0.7748 & 0.7990 & 0.9401 & 2.1963 & 0.7267 & 0.9108 \\
MBMF \cite{gong2016motion} & 26.59 & 0.7418 & 0.8079 & 0.8741 & 0.9576 & 2.2407 & 0.7421 & 0.9221 \\
MS-CNN \cite{nah2016deep} & 26.51 & 0.7432 & 0.8083 & 0.8481 & 0.9587 & 2.2235 & 0.7298 & 0.9224 \\
\hline
Ours (final) & 27.08 & \textbf{0.7510} & \textbf{0.8120} & 0.8743 & \textbf{0.9651} & \textbf{2.5192} & \textbf{0.7718} & \textbf{0.9318}\\
\hline
\end{tabular}
\caption{Quantitative Comparison of our method with other state of the art blind deblurring algorithms on K\"{o}hler Dataset.\label{Table5}}
\end{table*}

\begin{figure*}[htbp]
\centering 
\subfigure[Blurry Input Image]{\includegraphics[width= 0.25\textwidth]{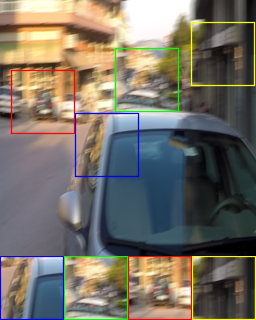}}%
\subfigure[Ground Truth]{\includegraphics[width= 0.25\textwidth]{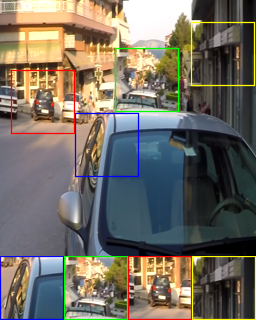}}%
\subfigure[Xu et al.\cite{xu2013unnatural}]{\includegraphics[width= 0.25\textwidth]{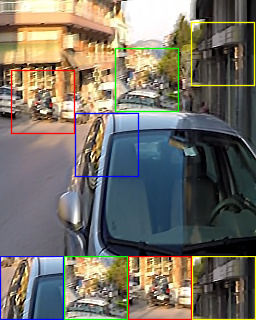}}%
\subfigure[Whyte \cite{whyte2014deblurring}]{\includegraphics[width= 0.25\textwidth]{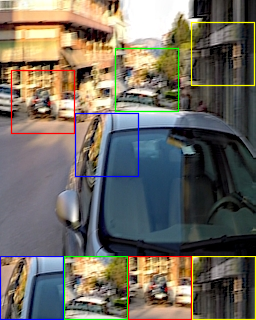}} \\
\subfigure[Sun et al. \cite{sun2015learning}]{\includegraphics[width= 0.25\textwidth]{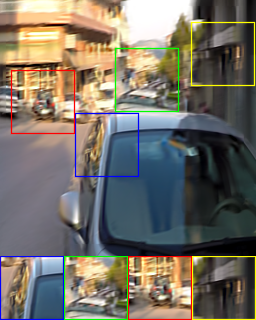}}%
\subfigure[MBMF \cite{gong2016motion}]{\includegraphics[width= 0.25\textwidth]{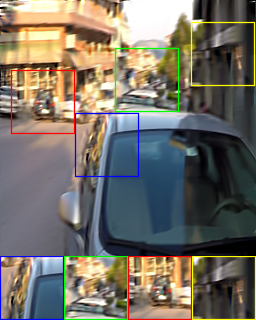}}%
\subfigure[MS-CNN \cite{nah2016deep}]{\includegraphics[width= 0.25\textwidth]{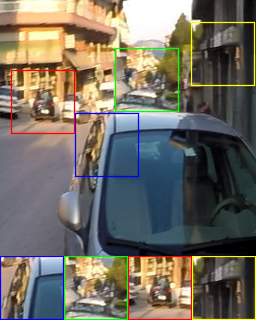}}%
\subfigure[OURS]{\includegraphics[width= 0.25\textwidth]{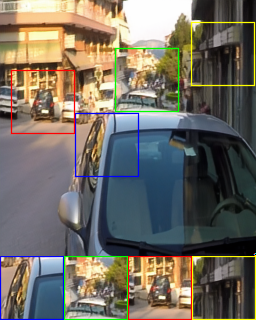}} \\
\caption{Comparison of deblurred images by our model and other algorihtms on one of the images taken from GoPro dataset \cite{nah2016deep}.  \label{figure4}}
\end{figure*}
\begin{figure*}[htbp]
\centering 
\subfigure[Blurry Input Image]{\includegraphics[width= 0.25\textwidth]{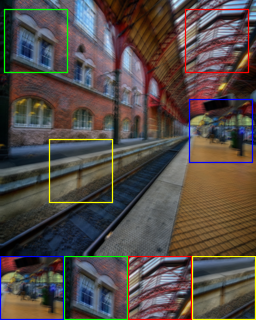}}%
\subfigure[Ground Truth]{\includegraphics[width= 0.25\textwidth]{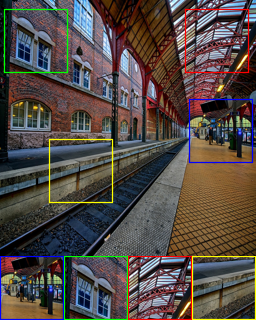}}%
\subfigure[Xu et al.\cite{xu2013unnatural}]{\includegraphics[width= 0.25\textwidth]{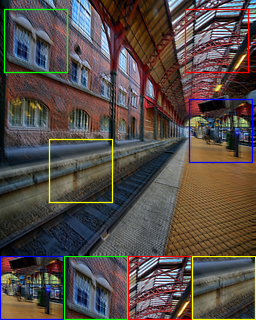}}%
\subfigure[Whyte \cite{whyte2014deblurring}]{\includegraphics[width= 0.25\textwidth]{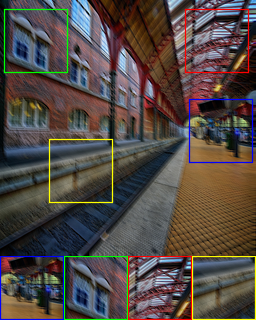}} \\
\subfigure[Sun et al. \cite{sun2015learning}]{\includegraphics[width= 0.25\textwidth]{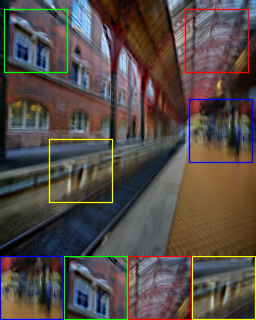}}%
\subfigure[MBMF \cite{gong2016motion}]{\includegraphics[width= 0.25\textwidth]{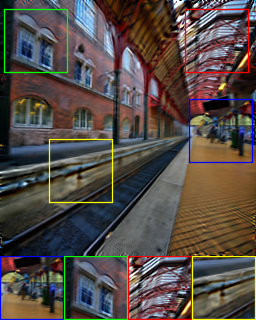}}%
\subfigure[MS-CNN \cite{nah2016deep}]{\includegraphics[width= 0.25\textwidth]{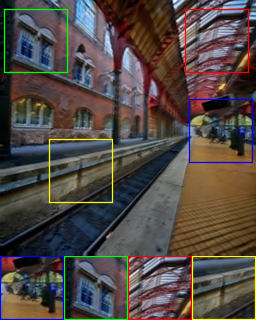}}%
\subfigure[OURS]{\includegraphics[width= 0.25\textwidth]{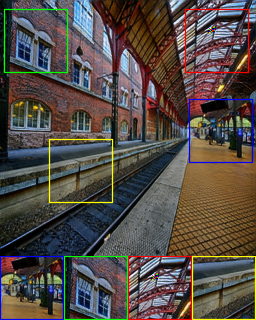}} \\
\caption{Comparison of deblurred images by our model and other algorihtms on one of the images taken from Lai et al.'s Dataset \cite{lai2016comparative}. \label{figure2}}
\end{figure*}
\subsection{Model Architecture}
Our architecture consists of a densely connected generator and a discriminator. The task of the generator is to recycle features spanning across multiple receptive scales to generate an image that fools the discriminator into thinking that the generated image came from the target distribution. Thus, we can generate visually appealing and statistically consistent deblurred image given a blurred image. The task of the discriminator is to correctly identify from which distribution each of its input images came from by analysing different patches in each image to make a decision. We elaborate both our generator and discriminator models in detail.
\subsubsection{The Generator} 
Unlike \cite{huang2016densely}, we do not reduce the dimension of the information and keep it constant throughout the network. While this does give rise to memory constraints, it protects the network from generating checkerboard artifacts found commonly in networks relying on deconvolution to generate visually appealing images \cite{isola2016image}. Instead, through feature re-use across all levels in the generator network, our model exhibits high generation performance with a much smaller network depth than the other CNN-based methods used for non-uniform motion deblurring \cite{nah2016deep},\cite{gong2016motion},\cite{sun2015learning}. This enables smoother training, faster test time and allows efficient memory usage. Our generator model as shown in Fig. \ref{CNN} consists of 4 parts which are the head, the dense field, the tail, and the global skip connection. We describe each of them in detail below. \\
\textbf{a) The Head:} We define the hyper-parameter `channel-rate' (chr) as the constant number of activation channels that are output by each convolutional layer. The value of channel-rate is 64. The head comprises of a simple $3\times 3$ convolutional layer which convolves over the raw input image and outputs $4\times $channel-rate (256) feature activations. This provides sufficient first-level activation maps to trigger the densely connected stack of layers.\\
\textbf{b) The Dense Field:} This section consists of $N$ number of convolutional `blocks' placed sequentially one after the other, all having their outputs fully connected with the output of the layer ahead of them. The dense connection is efficiently achieved in practice by concatenating output activation maps of every $i$th layer  in the dense field with the output maps of $(i+1)$th layer. Hence, the number of activation maps input to the $m$th dense block will be equal to `$4\times$chr $+$ $(m-1)\times$chr'. The structure of a dense block is shown in Fig. \ref{denseblock}. The first operation is a Leaky ReLU \cite{maas2013rectifier} which not only adds non-linearity to the incoming activations but also avoids using sparse gradients which could compromize GAN training stability. The $1\times 1$ convolution `chokes' the number of activation maps being convolved later to a maximum equal to `$4\times$chr'. This conserves parameter and data memory in the deeper layers of the dense field when the number of raw activation channels entering will be $6\times$chr (384) or more. The convolution at the final layer of each dense block uses `chr' number of $3\times 3\times (4\times$chr) filters, giving rise to `chr' number of activation maps at the end of each dense block. The $3\times 3$ convolutions along the dense field are alternated between `spatial' convolution and `dilated' convolution with linearly increasing dilation factor \cite{yu2015multi}. We use dilated convolution \cite{yu2015multi} at every even numbered layer within the dense field. We have the dilation factor increasing linearly to a maximum till the centre of the dense field and then symmetrically reducing till we arrive at the tail. This helps to increase the receptive field at an exponential rate with every layer while the parameter space increases linearly and hence introduces higher disparity between the multiple scales of activation maps that arrive at subsequent dense layers. We avoid pooling and strided convolution operations to keep the dimensions of the output maps to be constant and equal to the image size throughout the network. Adding dropout at the end of each block helps us effectively add Gaussian noise to the input of each layer in the  generator (G) which prevents the GAN collapse problem by enabling G to blindly model shake distributions other than a pure delta distribution.  \\
\textbf{c) The Tail:} The Tail adds the non-linearity and through $1\times 1$ convolution increases the number of feature maps to $4\times$chr.\\
\textbf{d) The Global Skip Connection:} Deep generative CNNs usually face the problem of often inadvertently memorizing high level representations of edges as it is non-trivial to generalize over first-level features using several convolution operations. This would lead the network to not be able to retrieve sharp boundaries at correct locations from the shaken images. We concancate the output from the head of the network with the output of the tail. This gives rise to a good improvement in generation performance because the gradients can now flow from the tail straight to the first level convolutional layer and impact the update in the lower layers \cite{he2016identity}. But more importantly, this single connection `drives' the entire dense field in the centre to expend its `full knowledge' of the image towards understanding the residual between the ground truth and the blurred images. Meanwhile, it also optimizes gabor-like features
of our CNN directly from the ground truth fed into the generator-end \cite{zeiler2014visualizing}.
However, different from the traditional residual networks used in image restoration models, we do not use cascaded skip summations. Instead, we pass lower level knowledge to the upper layers through dense connection and direct the entire dense field to solely calculate the global residual, which as experiments show, enable our network to learn faster, achieve better convergence and show significantly better deblurring performance. 
\subsubsection{The Discriminator} In our GAN framework, the discriminator is the primary agent which guides the statistics that the generator employs to create restored images. Moreover, we do not want the depth of the discriminator network depth so much that it memorizes the easier task of classification. We employ a Markovian patch discriminator \cite{li2016precomputed} with 10 convolutional layers, which is similar to a non-overlapping sliding window that tends to look for well-defined, structural features at several local patches. This also enforces rich coloration in the generated natural images \cite{isola2016image}.
\subsection{Loss Functions}
a) $\ell_{1}$ \textbf{and Adverserial Loss}: Traditionally, learning-based image restoration works have used $\ell_{1}$ or $\ell_{2}$ loss between the ground truth and the rectified image as the chief objective function \cite{bruna2015super}. In case of an adverserial framework used for such a purpose \cite{nah2016deep}, this loss is pooled with the adverserial loss which measures how well the generator is performing with respect to fooling the discriminator. However, using $\ell_{1}$loss solely in deep CNN models leads to overly smooth images, as pixel-wise error functions tend to converge at the mean of all possible solutions in the image manifold, whenever they encounter uncertainty \cite{bruna2015super}. This creates dull images with not many sharp edges and most importantly, with the blur still largely intact at edges and corners. At the same time, solely using adverserial loss does retain edges and gives rise to a more realistic color distribution \cite{bruna2015super}. However, it compromises on two things: it still has no abstract idea of structure and it only has the discriminator judging generator performance based on the output image alone with no regard to the blurred input. We remove these limitations by leveraging perceptual loss and adding it to the net loss function given in Eqn. \ref{netlossfunction}.\\
b) \textbf{Perceptual Loss:} We need to augment structural knowledge into the generator to counter the patch-wise judgement of the Markovian discriminator. One such loss function, as introduced in \cite{johnson2016perceptual} is the Euclidean difference between deep convolutional activations of the ground truth and generated latent image which is also known as `perceptual loss'. This loss term is given in Eqn. \ref{Equation1},  
\begin{equation} \label{Equation1}
\begin{split}
\mathcal{L}_{percep (VGG/i,j)} = \frac{1}{W_{i,j}H_{i,j}} \sum_{x=1}^{W_{i,j}} \sum_{y=1}^{H_{i,j}} (\phi_{i,j}(I^{\text{Groundtruth}})_{x,y}-  \\
 \phi_{i,j}(G_{\theta_{G}}(I^{\text{Blurred}}))_{x,y})^{2}         
\end{split}
\end{equation}
Here, $W_{i,j}$, $H_{i,j}$ are the width and height of the $(i,j)^{th}$ ReLU layer of VGG-16 network \cite{simonyan2014very} and $\phi_{i,j}$ is the forward pass through VGG-16 network upto ReLU $3\_ 3$ layer.

\subsubsection{Conditional adversarial framework} We feed two image pairs into the discriminator in our GAN framework. One pair consists of the input blurred image and the corresponding output image generated by the generator, whereas the other pair consists of the input blurred image and the corresponding ground truth deblurred image. This converges with the generator modelling the conditional distribution of the latent image, given the input image, a result that will help the generated images maintain high statistical consistency between a given input and its output. This is essentially what we need, because we want `G' to maintain the output's dependency on the blurred input to accomodate different kinds and amounts of shake blur and prevent it from swaying too far away in its effort to fool the discriminator. Hence, we can view a conditional GAN as a `relevance regularizer' in an image to image network. 
Mathematically, this would change the original GAN optimization problem used in our task which would be given by:
\begin{equation}
\begin{split}
\underset{\theta_{G}}{\mathrm{ min}}\hspace{0.1cm}\underset{\theta_{D}}{\mathrm{ max}}\hspace{0.1cm} {\mathbb{E}}_{I^{\text{Groundtruth}} \sim p_{\text{train}}(I^{\text{Groundtruth}})} [\text{log}\hspace{0.05cm}D_{\theta_{D}}(I^{\text{Groundtruth}})] + \\
 {\mathbb{E}}_{I^{\text{Groundtruth}} \sim p_{\text{train}}(I^{\text{Groundtruth}})} [\text{log}\hspace{0.05cm}(1 - D_{\theta_{D}} (G_{\theta_{G}} (I^{\text{Blurred}}))]
\end{split}
\end{equation} 
to a conditional loss function which needs to be minimized, given by
\begin{equation}
\mathcal{L}_{\text{ConditionalGAN}}^{\text{Generator}} = -{\mathbb{E}}_{I \epsilon (I^{\text{Blurred}})} \big [\text{log} D_{\theta_{D}} (G_{\theta_{G}}(I^{\text{Blurred}})|I^{\text{Blurred}})]
\end{equation}
Thus, the combined loss function for our network is,
\begin{equation} \label{netlossfunction}
\mathcal{L}_{net} = \mathcal{L}_{\text{ConditionalGAN}}^{\text{Generator}} + (K_{1})\mathcal{L}_{percep} + (K_{2})\mathcal{L}_{L_{1}}
\end{equation}
where, $K_{1}$ and $K_{2}$ are hypermeters which are set to 145 and 170 respectively in our experiments. From Table \ref{Table2}, we notice a significant boost in the performance across all metrics by introducing this technique. At this stage, our network has already outperformed the two baseline models modified and trained for our task: a very-deep, sequential ResNet model used by \cite{ledig2016photo} and the hourglass, U-net model used by \cite{isola2016image}. It is worth noting that our dense model with much fewer layers (10 dense blocks) not only outperformed, but also converged faster than the model in \cite{ledig2016photo} with 15 residual blocks, showing that our model and the framework resonate much better. 
\section{Experiments}

\subsection{Experimental Settings}
We implemented our model with torch7 library. All the experiments were performed on a workstation with i7 processor and NVIDIA GTX Titan X GPU.\\
\textbf{Network Parameters:} We optimize our loss function through the ADAM scheme \cite{kingma2014adam} and converge it using stochastic gradient descent (SGD). Throughout the experiments, we kept the batch-size for training as 3 and fixed base learning rate and momentum to $0.0002$ and $0.9$ respectively. Similar to \cite{isola2016image}, we use instance normalization instead of training batch statistics during test-time. \\
\textbf{Experiments for further architectural considerations} : We also perform a simple ablation study over the architecture of our fully evolved model to isolate which connections in the dense net are more important towards image restoration to further explore our own model. We use two sub-dense architecures named `A' and `B' to do so. The results of the ablation studies are given in the Table \ref{Table1}.\\
\textbf{A)} In this model, the three lower and higher extreme layers of our `dense field' are replaced with successive residuals of \cite{ledig2016photo} and \cite{nah2016deep} while the middle layers remain dense. We noticed a significant drop in performance compared to our final model by doing so. This is because the central part has `forgotten' entry-level features which were crucial in calculating the global residual between the head and the tail. \\
\textbf{B)} Switching the locations of the residuals and the dense layers leads to better performance than having both a fully residual network \cite{ledig2016photo} and a centrally dense network (A). Although it is slightly outperformed by our final model, it saves a dramatic amount of GPU memory by cutting down a lot of data concatenation. Hence, dense connections work best when connections between the farthest of layers is achieved. This helps the network to keep recycling lower features for globalizing the knowledge of the higher layers.

\subsection{Datasets}
To train our model, we extracted patches of size $256 \times 256 \times 3$ from GoPRO dataset and combined them with the images sampled randomly from MS-COCO \cite{lin2014microsoft} and Imagenet dataset \cite{deng2009imagenet} (which are resized to $ 256 \times 256 \times 3$) to generate our training dataset. We then apply non-uniform blurs similar to \cite{lai2016comparative} on images sampled from MS-COCO and ImageNet datasets. We also perform data augmentation by using translational and rotational flipping, thus producing a final dataset consisting of 0.5 million training image pairs of blurred and deblurred images. \\
We perform comparison of progressive models on one dataset generated synthetically by us and compare the performance of our method with the other state-of-the-art methods on three different benchmark datasets. Following Lai et al. \cite{lai2016comparative}, we used eight full reference metrics for quantitative analysis of our deshaking model. Detailed descriptions of these metrics can be found in \cite{lai2016comparative}. For comparison we choose the state-of-the-art blind deblurring algorithms given by: Xu et al. \cite{xu2013unnatural}, Whyte \cite{whyte2014deblurring}, Sun et al. \cite{sun2015learning}, MBMF \cite{gong2016motion}, and MS-CNN \cite{nah2016deep}.\\
a) \textbf{Places Dataset \cite{zhou2014learning}:} To perform the quantitative comparison of progressive models, we generate a synthetically blurred dataset in the same way as described earlier. We used the images from the Places dataset to generate pairs of deblurred and blurred images. The results are shown in Table \ref{Table2}. Note that Ours(1) in the Table \ref{Table2} describes the dense generative net with only the $\ell_{1}$ loss.\\
b) \textbf{GoPro Dataset} \cite{nah2016deep}: Images in this dataset were captured using GoPro and closely mimic the blur generated in real life. Out of total images, we used 438 images for our testing dataset and the rest of the images for creating the training dataset. We show the results of the quantitative comparison with the other state-of-the-art methods in Table \ref{Table3}. Our results show significant improvement in terms of image quality. \\
c) \textbf{Lai et al. Dataset} \cite{lai2016comparative}: Lai et al. generated synthetic dataset by convolving nonuniform blur kernels and imposing several common degradations. To generate blur kernels they also recorded 6D camera trajectories. The comparative methods of our method with other algorithms are given in Table \ref{Table4}. The MS-CNN learning model \cite{nah2016deep}, which also bypasses the kernel estimation step, was re-trained by us on the same dataset that we used for training our own model. On the other hand, we use the available testing codes of \cite{sun2015learning} and \cite{gong2016motion} for reporting the comparison.  \\
d) \textbf{K\"{o}hler et al. Dataset} \cite{kohler2012recording}: This benchmark dataset consists of four latent images. To construct a non-uniform blur dataset, twelve 6D camera trajectories were recorded over time assuming linear camera response function using which blurred images were captured. The captured scenes were planar and at a fixed distance from the camera. We report the quantitative results on this dataset in Table \ref{Table5}. From the table, we could infer that our model significantly outperforms the other methods.
Qualitative comparisons of the different methods with ours could be seen in Fig. \ref{figure1}, Fig. \ref{figure4} and Fig. \ref{figure2}. As evident from the figures, results produced by our method are visually superior compared to that of the-state-of-the-art.
\section{Conclusion}
We have designed a novel, end-to-end conditional GAN-based filter model which performs blind restoration of shaken images. Our results show that our model and framework outperforms the state-of-the-art for non-uniform deblurring. The fast execution time of our model makes it easily deployable in cameras and photo editing tools. We show that densely connected convolutional networks can be as effective for image generation as it is for classification. 
\section*{Acknowledgement}
Shubham Pachori and Shanmuganathan Raman were supported through an ISRO RESPOND grant.  

\bibliographystyle{ieee}

\begin{thebibliography}{10}\itemsep=-1pt

\bibitem{bruna2015super}
J.~Bruna, P.~Sprechmann, and Y.~LeCun.
\newblock Super-resolution with deep convolutional sufficient statistics.
\newblock In {\em ICLR}, 2016.

\bibitem{deng2009imagenet}
J.~Deng, W.~Dong, R.~Socher, L.-J. Li, K.~Li, and L.~Fei-Fei.
\newblock Imagenet: A large-scale hierarchical image database.
\newblock In {\em CVPR 2009. IEEE Conference on}, pages 248--255. IEEE, 2009.

\bibitem{gong2016motion}
D.~Gong, J.~Yang, L.~Liu, Y.~Zhang, I.~Reid, C.~Shen, A.~v.~d. Hengel, and
  Q.~Shi.
\newblock From motion blur to motion flow: a deep learning solution for
  removing heterogeneous motion blur.
\newblock In {\em IEEE CVPR}, 2017.

\bibitem{harmeling2010space}
S.~Harmeling, H.~Michael, and B.~Sch{\"o}lkopf.
\newblock Space-variant single-image blind deconvolution for removing camera
  shake.
\newblock In {\em Advances in NIPS}, pages 829--837, 2010.

\bibitem{he2016identity}
K.~He, X.~Zhang, S.~Ren, and J.~Sun.
\newblock Identity mappings in deep residual networks.
\newblock In {\em ECCV}, pages 630--645. Springer, 2016.

\bibitem{huang2016densely}
G.~Huang, Z.~Liu, K.~Q. Weinberger, and L.~van~der Maaten.
\newblock Densely connected convolutional networks.
\newblock In {\em IEEE CVPR}, 2017.

\bibitem{isola2016image}
P.~Isola, J.-Y. Zhu, T.~Zhou, and A.~A. Efros.
\newblock Image-to-image translation with conditional adversarial networks.
\newblock In {\em IEEE CVPR}, 2017.

\bibitem{johnson2016perceptual}
J.~Johnson, A.~Alahi, and L.~Fei-Fei.
\newblock Perceptual losses for real-time style transfer and super-resolution.
\newblock In {\em ECCV}, pages 694--711. Springer, 2016.

\bibitem{kingma2014adam}
D.~Kingma and J.~Ba.
\newblock Adam: A method for stochastic optimization.
\newblock In {\em ICLR}, 2015.

\bibitem{kohler2012recording}
R.~K{\"o}hler, M.~Hirsch, B.~Mohler, B.~Sch{\"o}lkopf, and S.~Harmeling.
\newblock Recording and playback of camera shake: Benchmarking blind
  deconvolution with a real-world database.
\newblock {\em Computer Vision--ECCV 2012}, pages 27--40, 2012.

\bibitem{lai2016comparative}
W.-S. Lai, J.-B. Huang, Z.~Hu, N.~Ahuja, and M.-H. Yang.
\newblock A comparative study for single image blind deblurring.
\newblock In {\em CVPR}, pages 1701--1709, 2016.

\bibitem{ledig2016photo}
C.~Ledig, L.~Theis, F.~Husz{\'a}r, J.~Caballero, A.~Cunningham, A.~Acosta,
  A.~Aitken, A.~Tejani, J.~Totz, Z.~Wang, et~al.
\newblock Photo-realistic single image super-resolution using a generative
  adversarial network.
\newblock In {\em IEEE CVPR}, 2017.

\bibitem{li2016precomputed}
C.~Li and M.~Wand.
\newblock Precomputed real-time texture synthesis with markovian generative
  adversarial networks.
\newblock In {\em ECCV}, pages 702--716. Springer, 2016.

\bibitem{lin2014microsoft}
T.-Y. Lin, M.~Maire, S.~Belongie, J.~Hays, P.~Perona, D.~Ramanan,
  P.~Doll{\'a}r, and C.~L. Zitnick.
\newblock Microsoft coco: Common objects in context.
\newblock In {\em ECCV}, pages 740--755. Springer, 2014.

\bibitem{maas2013rectifier}
A.~L. Maas, A.~Y. Hannun, and A.~Y. Ng.
\newblock Rectifier nonlinearities improve neural network acoustic models.
\newblock In {\em ICML}, volume~30, 2013.

\bibitem{nah2016deep}
S.~Nah, T.~H. Kim, and K.~M. Lee.
\newblock Deep multi-scale convolutional neural network for dynamic scene
  deblurring.
\newblock In {\em IEEE CVPR}, 2017.

\bibitem{simonyan2014very}
K.~Simonyan and A.~Zisserman.
\newblock Very deep convolutional networks for large-scale image recognition.
\newblock In {\em ICLR}, 2015.

\bibitem{sun2015learning}
J.~Sun, W.~Cao, Z.~Xu, and J.~Ponce.
\newblock Learning a convolutional neural network for non-uniform motion blur
  removal.
\newblock In {\em IEEE CVPR}, pages 769--777, 2015.

\bibitem{vasiljevic2016examining}
I.~Vasiljevic, A.~Chakrabarti, and G.~Shakhnarovich.
\newblock Examining the impact of blur on recognition by convolutional
  networks.
\newblock {\em arXiv preprint arXiv:1611.05760}, 2016.

\bibitem{whyte2014deblurring}
O.~Whyte, J.~Sivic, and A.~Zisserman.
\newblock Deblurring shaken and partially saturated images.
\newblock {\em IJCV}, 110(2):185--201, 2014.

\bibitem{xu2013unnatural}
L.~Xu, S.~Zheng, and J.~Jia.
\newblock Unnatural l0 sparse representation for natural image deblurring.
\newblock In {\em CVPR}, pages 1107--1114, 2013.

\bibitem{yu2015multi}
F.~Yu and V.~Koltun.
\newblock Multi-scale context aggregation by dilated convolutions.
\newblock In {\em ICLR}, 2016.

\bibitem{zeiler2014visualizing}
M.~D. Zeiler and R.~Fergus.
\newblock Visualizing and understanding convolutional networks.
\newblock In {\em ECCV}, pages 818--833. Springer, 2014.

\bibitem{zhou2014learning}
B.~Zhou, A.~Lapedriza, J.~Xiao, A.~Torralba, and A.~Oliva.
\newblock Learning deep features for scene recognition using places database.
\newblock In {\em NIPS}, pages 487--495, 2014.

\end{thebibliography}

\end{document}